\documentclass[runningheads]{llncs}
\usepackage{graphicx}
\usepackage{multirow}
\usepackage{amssymb}
\usepackage{array}
\usepackage{cite}
\usepackage{amsmath}
\usepackage{soul}
\usepackage{xcolor}
\usepackage{hyperref}

\newcolumntype{C}[1]{>{\centering\let\newline\\\arraybackslash\hspace{0pt}}m{#1}}
\newcolumntype{L}[1]{>{\raggedright\let\newline\\\arraybackslash\hspace{0pt}}m{#1}}
\begin{document}
\title{Few Is Enough: Task-Augmented Active Meta-Learning for Brain Cell Classification}
% \title{Fast adaptation: Task-augmented meta-learning used in brain cell classification problems\thanks{Supported by organization x.}}
\author{Pengyu Yuan\inst{1} \and Aryan Mobiny\inst{1} \and Jahandar Jahanipour\inst{1,2} \and Xiaoyang Li\inst{1} \and Pietro Antonio Cicalese\inst{1} \and Badrinath Roysam\inst{1} \and Vishal Patel\inst{3} \and Maric Dragan\inst{2} \and Hien Van Nguyen\inst{1}}
%
%index{Yuan, Pengyu}
%index{Mobiny, Aryan}
%index{Jahanipour, Jahandar} 
%index{Li, Xiaoyang} 
%index{Cicalese, Pietro Antonio}
%index{Roysam, Badrinath} 
%index{Patel, Vishal} 
%index{Dragan, Maric} 
%index{Van Nguyen, Hien} 
%
\authorrunning{P. Yuan et al.}

\institute{University of Houston, Houston, Texas, USA \\ \email{pyuan2@uh.edu} \and
National Institutes of Health, Bethesda, Maryland, USA \and
Johns Hopkins University, Baltimore, Maryland, USA }
% \email{***@***.***}}
% \institute{University of Houston 
% \email{pyuan2@uh.edu}}
\maketitle              % typeset the header of the contribution
\begin{abstract}

Deep Neural Networks (or DNNs) must constantly cope with distribution changes in the input data when the task of interest or the data collection protocol changes. Retraining a network from scratch to combat this issue poses a significant cost. Meta-learning aims to deliver an adaptive model that is sensitive to these underlying distribution changes, but requires many tasks during the meta-training process. In this paper, we propose a tAsk-auGmented actIve meta-LEarning (AGILE) method to efficiently adapt DNNs to new tasks by using a small number of training examples. AGILE combines a meta-learning algorithm with a novel task augmentation technique which we use to generate an initial adaptive model. It then uses Bayesian dropout uncertainty estimates to actively select the most difficult samples when updating the model to a new task. This allows AGILE to learn with fewer tasks and a few informative samples, achieving high performance with a limited dataset. We perform our experiments using the brain cell classification task and compare the results to a plain meta-learning model trained from scratch. We show that the proposed task-augmented meta-learning framework can learn to classify new cell types after a single gradient step with a limited number of training samples. We show that active learning with Bayesian uncertainty can further improve the performance when the number of training samples is extremely small. Using only 1\% of the training data and a single update step, we achieved 90\% accuracy on the new cell type classification task, a 50\% points improvement over a state-of-the-art meta-learning algorithm.

% This requirement limits the applicability of this technology because in many cases users only have limited source data but desire the networks to be able to adapt efficiently when there's a change in the data.

% \vspace{-3mm}
\end{abstract}
%
%
%
% \vspace{-3mm}
\section{Introduction}
% \vspace{-3mm}
% % PROBLEM OF FEW TRAINING SAMPLES
% Supervised learning has shown tremendous ability in machine learning and has been successfully applied in many areas especially in computer vision. However, supervised learning especially in deep learning requires a large amount of labeled data for good generalization. For example, MNIST dataset enables handwriting recognition and Imagenet unleashed the potential of deep learning on natural image classification. Data acquisition is not an easy task for many important applications such as cell microscopy and seismic image processing. Besides that, labeling data can be expensive or time-consuming so that the deep learning algorithm requires less training data will always be preferred. 

% METHODS/TOPICS TO ADDRESS FEW TRAINING SAMPLES PROBLEM

The ability of Deep Neural Networks (or DNNs) to generalize to a given target concept is dependent on the amount of training data used to generate the model. This is a significant limitation, as many real-world classification tasks depend on a limited number of training samples for accurate classification. Various research groups have developed techniques that utilize different underlying principles to address this issue. These techniques include data augmentation\cite{shorten2019survey} and generative models \cite{zhu2009introduction, kingma2014semi, goodfellow2014generative}, which try to directly increase the number of training samples. Active learning is another technique which aims to select the most valuable samples to include in the training set\cite{settles2009active, sener2017active}. Finally, transfer learning allows the model to adapt to new and unseen tasks by using its own previous knowledge, often reducing the amount of data needed for generalization\cite{bengio2012deep, gopalan2011domain, saenko2010adapting, saenko2010adapting}. 
% This technique depends on the transformation of data, features\cite{gopalan2011domain} or the classifier itself\cite{saenko2010adapting} to minimize the effects caused by the perturbation in the data distribution.

% LIMITATION OF TRANSFER LEARNING/META LEARNING IN MEDICAL IMAGE PROCESSING
Medical data is characterized by significant distribution shifts and small samples sets that negatively affect the quality of the generated model. In the cell classification task\cite{zeng2017neuronal}, there are several contributing factors for poor generalization; different biomarkers, unique cell morphologies, variations in stain intensity and image quality all contribute to the variability of the data. Each of these factors could be considered a unique parameter with which to create unique classification tasks. Traditional transfer learning methods pre-trained on various source tasks may not perform well; poor model initialization parameters coupled with unadjusted hyperparameters may cause the model to fall into a bad local minima. Mainstream transfer learning methods also require a time-consuming model retraining process \cite{pan2009survey}. In recent years, a more advanced model architecture called meta-learning has been developed to address these adaptability issues\cite{vilalta2002perspective}. Meta-learning approaches try to generate a more robust model that can learn to quickly adapt to new tasks with minimal labeled samples. Although meta-learning does not require many labeled samples for each task, it requires many different tasks to effectively learn how to adapt; this may be a problem when the number of tasks is limited. This technique also selects training samples randomly for each new task, which may negatively impact the strength of the model if it is trained on easier samples.

% PROPOSE TASK-AUGMENTED META-LEARNING + ACTIVE LEARNING WITH BAYESIAN DROPOUT UNCERTAINTY MECHANISM
In this paper, we utilize various strategies to create an adaptive framework called tAsk-auGmented actIve meta-LEarning (or AGILE) which allows the classifier to achieve high performance with few training samples and gradient updates for each new task. The experiments we perform on the brain cell type classification task show that the AGILE classifier can quickly adapt to new cell types by utilizing very few labeled training samples.

% \vspace{-3mm}
\section{Related works}
% \vspace{-2mm}
% BRAIN CELL TYPE CLASSIFICATION
\subsubsection{Brain cell type classification}
The brain is a highly complex organ made up of myriad different cell types, each with their own unique properties \cite{cellofthebrain}. Gene expression experiments have highlighted cell type composition based on the expression value of markers for five major cell types: neurons, astrocytes, oligodendrocytes, microglia, and endothelial cells \cite{de2014alzheimer, mckenzie2018brain}. Besides that, there are around 50-250 neuronal sub-cell types purported exist \cite{ascoli2007neuromorpho}. Different cell types may express different biomarkers or unique combinations of biomarkers, with some of then shared with other cell types. Correctly identifying the cell type using these biomarkers is essential for many medical researches such as schizophrenia \cite{skene2018genetic} and brain cell type specific gene expression \cite{mckenzie2018brain}. It is easy to train a deep neural network to identify one cell type but it is not easy to scale the network for hundreds of new cell types. Our meta-learning framework aims to provide an adaptive model that can rapidly adjust itself to new classification tasks.
%Meta learning can help us get an adaptive model which can easily adapt to new classification tasks.

% META-LEARNING
% \vspace{-3mm}
\subsubsection{Meta-learning}
% \vspace{-3mm}
% Real-life applications require the ability of quickly adapt from few samples for unseen task. 
Meta-learning aims to study how meta-learning algorithms can acquire fast adaptation capability from a collection of tasks \cite{bengio1990learning, bengio1992optimization, vilalta2002perspective, santoro2016one}. Meta-learning often consists of a meta-learner and a learner that learn at two levels of different time scales \cite{hochreiter1997long, schweighofer2003meta}. Santoro et al.  Koch et al. \cite{koch2015siamese}, Vinyals et al. \cite{vinyals2016matching} and Snell et al. \cite{snell2017prototypical} proposed to learn a robust kernel function of feature embeddings to illustrate the similarities between different samples. Another popular approach is to directly optimize the meta-learner through the gradient descent \cite{hochreiter2001learning, andrychowicz2016learning, ravi2016optimization, finn2017model}. Model agnostic meta-learning  (MAML)\cite{finn2017model} is proven to be one of the state-of-the-art approaches in the meta-learning field. These methods demonstrated human-level accuracy on many classification tasks. However most of them require a lot of tasks to train the meta-learner and it can not actively select training samples which might not be the optimal case in practice.

% Santoro et al. \cite{santoro2016one} trained a recurrent neural network on the entire dataset then used an augmented memory to store prominent features for each task.
\begin{figure}[!t]
\centering
\includegraphics[width=0.90\textwidth]{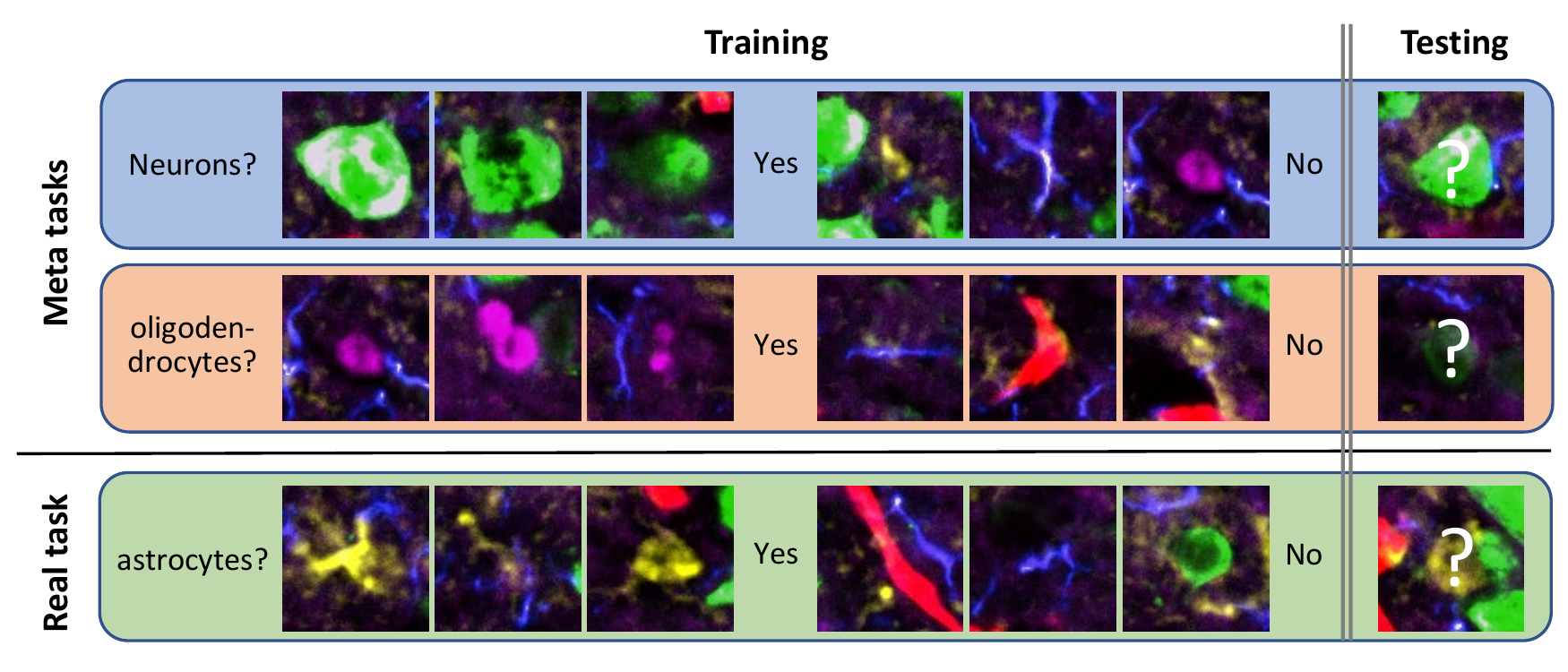}
% \vspace{-4mm}
\caption{Multi-task brain cell classification. Each task is a binary classification problem for a specific brain cell type. Cell to be classified is located in the center of the image. The model needs to be adapted to the unseen real task with few training samples.} 
% \vspace{-2mm}
\label{fig:tasks}
\end{figure}

% ACTIVE-LEARNING
% \vspace{-5mm}
\subsubsection{Active learning}
% \vspace{-3mm}
Active learning has been used to interactively and efficiently query information to achieve optimal performance for the task of interest. These methods select training samples based on information theory \cite{mackay1992information}, ensemble approaches \cite{mccallumzy1998employing}, and uncertainty measurements \cite{joshi2009multi, li2013adaptive}. However, these methods may not be effective for deep networks. Gal et al. \cite{gal2016dropout, gal2017deep}, Sener et al. \cite{sener2017active} and Fang et al. \cite{fang2017learning} did a lot of studies in finding heuristics of annotating new samples for deep networks. However, few of them have been applied to domain adaptation or meta-learning problems. Woodward et al. \cite{woodward2017active} added an active part in one-shot learning but did not utilize the best meta-learning structure. In this paper, we combine the advantages of both meta-learning and active-learning to get a fast adaptive model which can use the fewest data to achieve a good performance.

% Meta learning aims to present a model which can dynamically update itself according to the underlying changes in the data distribution.

% The plain transfer learning directly uses the model pre-trained on  $\{\mathcal{D}_i^\mathrm{train}\}_{i=1}^M$ where  $\mathcal{T}_i \in \mathcal{T}_\mathrm{meta}$, in all $M$ meta tasks as the initialization. It is not taught how to dynamically deal with data distribution shifts. Meta-learning, on the other hand, aims to learn the ability of fast adaptation from few labeled data samples by utilizing $\mathcal{D}_i^\mathrm{test}$ together with $\mathcal{D}_i^\mathrm{train}$ in the meta tasks.

\begin{figure}[!t]
\centering
\includegraphics[width=0.95\textwidth]{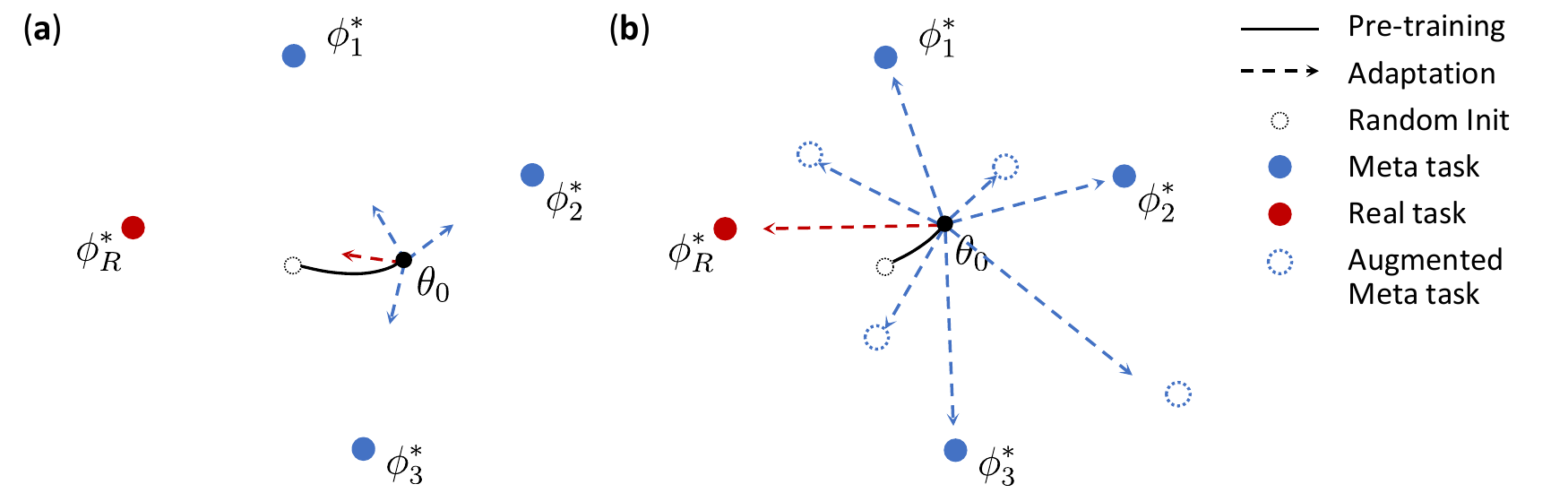}
% \vspace{-2mm}
\caption{Comparison of (\textbf{a})transfer learning and (\textbf{b})task-augmented meta-learning. (\textbf{a}) After pre-training on very few number of meta tasks, the model is pulled closer to the meta tasks and may be far from the real task. (\textbf{b}) By creating pseudo meta tasks to pre-train the meta-learning model, it gains the ability of adapting to the new tasks with only one gradient step updating.}
\label{fig:method}
% \vspace{-3mm}
\end{figure}

% \vspace{-3mm}
\section{Methodology}
% \vspace{-3mm}
% PROBLEM FORMULATION AND NOTATIONS
% \subsection{Problem formulation}
Consider a dataset consisting $Q$ samples:  $\mathcal{D}=\left\{\left(\mathbf{x}_{q}, \mathbf{y}_{q}\right)\right\}_{q=1}^{Q}$, where $(\mathbf{x}_q, \mathbf{y}_q)$ is an input-output pair sampled from the joint distribution $P(\mathcal{X}, \mathcal{Y})$. A task can be specifically defined by learning a model $f_{\phi}(\mathbf{x}):  \mathcal{X} \rightarrow \mathcal{Y}$ which is parameterized by $\phi$ to maximize the conditional probability $P_{\phi}(\mathcal{Y} | \mathcal{X})$. Thus whenever there is a change in the conditional distribution which is mainly caused by distribution shifts in $\mathcal{X}$ or $\mathcal{Y}$, it can be viewed as a new task. When dealing with mutiple tasks drawn from $P(\mathcal{T})$, each task $\mathcal{T}_i$ is associated with a unique dataset $\mathcal{D}_i$. We split these tasks into two parts: meta tasks $\mathcal{T}_\mathrm{meta}$ and real tasks $\mathcal{T}_\mathrm{real}$. Real tasks mean the model performance on these tasks is what we really care about and Meta tasks are what we used to pre-train the model. If there is no model adaptation, then the meta tasks are not needed. For each task $\mathcal{T}_i$, we have a train/test split $\mathcal{D}_i^\mathrm{train}/\mathcal{D}_i^\mathrm{test} \subset \mathcal{D}_i$ . The goal for task $\mathcal{T}_i$ is to learn a set of parameters $\phi_i$ from the $\mathcal{D}_i^\mathrm{train}$ to get the minimal loss on the test data, i.e. $\mathcal{L}(\phi_i, D_i^{test})$.

Our proposed framework AGILE has two phases, a task-augmented meta-learning method which learns to generate a strong model initialization which is sensitive to data distribution changes by using $\mathcal{T}_\mathrm{meta}$, and an active learning process which selects the most informative samples using Bayesian dropout uncertainties when apply the adaptive model on $\mathcal{T}_\mathrm{real}$. 
\subsection{Task-augmented meta-learning}
% \vspace{-2mm}
The phase I of our AGILE framework is the task-augmented meta-learning module. For the meta-learning setting, we have a learner which operates at fast time-scale and parameterized by $\phi \in \Phi$, and a meta-learner at a slower time-scale parameterized by $\theta \in \Theta$. The goal of meta-learning is to learn meta-parameters $\theta$ that can produce good task-specific parameters $\phi$ for all $M$ tasks after the fast adaptation:
% \vspace{-1mm}
\begin{equation}
% \vspace{-1mm}
    \theta^* = \underset{\theta \in \Theta}{\operatorname{argmin}}  \frac{1}{M} \sum_{i=1}^{M} \mathcal{L} \left ( \mathcal{A}dapt(\theta, \mathcal{D}_i^\mathrm{train}), \mathcal{D}_i^\mathrm{test}  \right )
    \label{eq:meta}
\end{equation}
where $\mathcal{A}dapt()$ function is an adaptation step completed by the learner. We employ model agnostic meta-learning (MAML)\cite{finn2017model} which initializes the model at $\theta$ then updates it using training data for each task $\mathcal{D}_i^\mathrm{train}$ as follows:
\begin{equation}
% \vspace{-1mm}
    \phi_i \equiv  \mathcal{A}dapt(\theta, \mathcal{D}_i^\mathrm{train}) = \theta \stackrel{Z}{-} \alpha \nabla_{\theta} \mathcal{L}(\theta, \left\{\left(\mathbf{x}_{k}, \mathbf{y}_{k}\right)\right\}_{k=1}^{K})), \quad \left(\mathbf{x}, \mathbf{y}\right) \sim \mathcal{D}_i^\mathrm{train}
    \label{eq:learner}
\end{equation}
where $\mathcal{L}$ is the loss function, $\alpha$ the learning rate for the learner, and $\stackrel{Z}{-}$ a short-hand notation for running a Z-step gradient descent which is relatively fast. A fixed number of K class-balanced samples $\left\{\left(\mathbf{x}_{k}, \mathbf{y}_{k}\right)\right\}_{k=1}^{K}$ randomly sampled from  $\mathcal{D}_i^\mathrm{train}$ are used to update the learner for every iteration.

% \vspace{-2mm}
\subsubsection{Task transformation}
% \vspace{-3mm}
Meta-learner is trained on the meta tasks. Our experiments show that when the number of meta tasks is not big enough, it can not obtain the general fast adaptation ability on $\mathcal{T}_\mathrm{real} \sim  P(\mathcal{T})$. Similar as data augmentation, we applied a set of task-augmentation functions $\{ \mathcal{G}_l \}_{l=1}^L: \mathcal{T} \rightarrow \mathcal{T}'$ on existing tasks to create new tasks. Because the task is specifically determined by the conditional probability $P(\mathcal{Y} | \mathcal{X})$, either change $\mathcal{X}$ or $\mathcal{Y}$ will lead to a new task. The task-augmentation functions we applied are (1) Flipping the label (2) Shuffling the order of input channels (3) Rotating the images. 
Considering the input image with $c$ biomarkers $\mathbf{x} \in \mathbb{R}^{w\times h \times c}$ and the binary label $y \in \mathbb{R}$, flipping labels is achieved by:
% \vspace{-1mm}
\begin{equation}
% \vspace{-1mm}
    y' = z(1 - y) + (1 - z)y, \quad \mathrm{where} \;  z \sim \mathrm{Bernoulli}(p_{f})
    \label{eq:flip}
\end{equation}
where $p_f$ is the probability of flipping the label. Shuffling the input channels is selected with a probability of $p_s$. By Constructing $c$ different one-by-one kernels $\left\{\mathbf{s}_{ij}\right\}_{i=1}^c \in \mathbb{R}^{1 \times 1 \times c}$ where $\mathbf{s}_{ij} = 1$ only if the $j^{th}$ bio-marker is placed at $i^{th}$ channel after the random shuffling, the shuffled images are obtained by the convolution:
% \vspace{-1mm}
\begin{equation}
% \vspace{-1mm}
    \mathbf{x'} = \mathbf{x} * \mathbf{s}_{ij}, \quad i, j = 1, 2, 3 \dots c  \
    \label{eq:shuffle}
\end{equation}

% use $c$ one by one convolutional kernels $\mathbf{s_j} \in \mathbb{R}^{1 \times 1 \times c}$ to select the $j^{th}$ bio-marker channel to put in the $i^{th}$ channel:
% \begin{equation}
%     \mathbf{x'} = \left\{\mathbf{x} * \mathbf{s_j}\right\}_{i=1}^c, \quad \mathbf{s_j} = 1 \; \
%     \label{eq:shuffle}
% \end{equation}

% Where $p_f$ is the probability of flipping the label. By constructing a shuffle matrix $\mathbf{S} \in \mathbb{R}^{c \times c}$ from the identical matrix $\mathbf{I} \in \mathbb{R}^{c \times c}$, where $s_{ij} = 1$ if the $j^{th}$ bio-marker is placed at $i^{th}$ channel after the random shuffling, we can get the new input:
% \begin{equation}
%     \mathbf{x'} = \mathbf{x} * \mathbf{S}, \quad \mathbf{S} = \mathbf{I} \;\mathrm{with} \; \mathrm{the}\; \mathrm{probability}\; \mathrm{of}\; (1-p_s)
%     \label{eq:shuffle}
% \end{equation}
% where $p_s$ is the probability of shuffling the input channel.

The third task augmentation method we used is to rotate the images for $90^{\circ}, 180^{\circ}\; \mathrm{or} \; 270^{\circ}$ with the probability of $p_r$. In the experiment, we set $p_f = p_s = p_r$. Theoretically any data augmentation techniques can be applied here to slightly change the distribution of $\mathcal{X}$. By increasing the diversity of the meta tasks, the meta-learner can be well trained to extract useful features after adapting to training data in any tasks. The comparison between task augmented meta-learning and plain transfer learning is illustrated in Figure \ref{fig:method}.

% ACTIVE LEARNING WITH UNCERTAINTY
% \vspace{-3mm}
\subsection{Active learning with Bayesian uncertainty}
% \vspace{-2mm}
The phase II of AGILE is to apply the pre-trained model with active learning. Previous meta-learning methods consider training and testing the meta-learning algorithm in the exactly same manner, which might not be practical. They fixed the number of training samples for each task. But for the real task $\{\mathcal{T}_\mathrm{real}\}_{j=1}^N$, this number may vary, especially when users want to annotate some new samples to improve the performance of the model. When the number of training samples is extremely small, how to unsupervisedly select the most informative samples to annotate is an important question. First, we use a random number of samples to update the learner during the meta-learning training process, so that the model is forced to learn with a different training size. 
% \vspace{-1mm}
\begin{equation}
% \vspace{-1mm}
    \phi_i = \theta \stackrel{Z}{-} \alpha \nabla_{\theta} \mathcal{L}(\theta, \left\{\left(\mathbf{x}_{k}, \mathbf{y}_{k}\right)\right\}_{k=1}^{\widetilde{K}})), \quad \left(\mathbf{x}, \mathbf{y}\right) \sim \mathcal{D}_i^\mathrm{train}
    \label{eq:learner}
\end{equation}
where $\widetilde{K}$ is varied between 1 sample from each class to the maximum number of samples $K$ allowed by the label budget. Second, we applied Bayesian dropout \cite{gal2016dropout, smith2018understanding} to get the uncertainties of the predictions over all unlabeled samples, which represents the current belief of the learner on the predictions:

% \vspace{-2mm}
\begin{equation}
% \vspace{-2mm}
    H\left(\mathbf{y} | \mathbf{x}, \mathcal{D}_j^\mathrm{train} \right)=-\sum_{\mathbf{y} \in \mathcal{Y}} p_{\mathrm{MC}}\left(\mathbf{y} | \mathbf{x}\right) \log p_{\mathrm{MC}}\left(\mathbf{y} | \mathbf{x}\right) 
    \label{eq:entropy}, \quad \mathrm{where} \; (\mathbf{x}, \mathbf{y}) \in \mathcal{D}_j^\mathrm{test} 
\end{equation}
$P_{\mathrm{MC}}\left(\mathbf{y} | \mathbf{x}\right)$ is the approximation of Bayesian inference with Monte Carlo integration $\frac{1}{T}\sum_{t=1}^{T} p_{ \phi_j} \left(\mathbf{y} | \mathbf{x} \right)$, where $p_{ \phi_j} \left(\mathbf{y} | \mathbf{x} \right)$ is the conditional probability of predicted class for input $\mathbf{x}$ in task $j$ and $T$ is the number of Monte-Carlo experiments. The higher entropy $H$ is, the higher uncertainty is in this prediction. For this Bayesian deep learning network, dropout is turned on both for the training and test time on $\mathcal{T}_\mathrm{real}$ with a drop rate of 0.1. Based on the uncertainties, the model will only select hard samples for training which significantly improves the performance.

\section{Experiments and Results}
% \vspace{-3mm}
For this brain cell type classification problem, we collected 4000 cells from 5 major cell types imaged from rat brain tissue sections: neurons, astrocytes, oligodendrocytes, microglia, and endothelial cells. There are 800 cells for each cell type. 7 biomarkers are used as the feature channels: DAPI, Histones, NeuN, S100, Olig 2, Iba1 and RECA1. DAPI and Histones are used to indicate the location of the cells while others are biomarkers for classification of specific cell types. Each cell is located in the center of a patch with a size of $100 \times 100$ pixels. As shown in Figure \ref{fig:tasks}, for each cell type, we have a binary classification task, 800 cells from this cell type and 800 from others. The model needs to correctly identify whether the given cell is this cell type. And it has no prior knowledge about the usage of each biomarker. Out of 1600 cells, 60\% of them are used as potential training samples and the rest are used for the test. Binary classification for neurons, oligodendrocytes and microglia are considered as the meta tasks. And the real tasks are to identify the remaining two cell types. 

% The model without the prior knowledge about the biomarker indication needs to correctly identify whether the given cell is this cell type.

% Elucidating brain cell type specific gene expression patterns is critical towards a better understanding
% of how cell-cell communications may influence brain functions and dysfunctions \cite{mckenzie2018brain}. It has been reported that approximately 50–250 neuronal sub-cell types exist \cite{ascoli2007neuromorpho, kandel2000principles, larson2013neurolex}, but the cell type composition is based on the expression value of markers for five major cell types: NeuN for pan-neuronal, IBA-1 for microglia, S100 for astrocytes, OLIG2 for oligodendrocytes, and RECA1 for endothelial cells. Two other biomarkers DAPI and Histones are also used to indicate the location of the cells. For each cell, we need to figure out the strongest protein expression to classify the cell so that all 7 biomarkers are necessary. 

% FAST ADAPTATION
% \subsection{Fast adaptation ability}
\begin{table}[!t]
    \caption{Methods configuration comparison which differ mainly in the data they use and the training framework. Meta-learning methods are supposed to perform well with few training samples and little training time.  (\# means the number of)}
    % \vspace{-2mm}
    % Meta-learning methods are supposed to perform well with few training samples and little training time for the real tasks.  (\# means the number of)}
    \centering
    \resizebox{0.9\columnwidth}{!}{%
    \begin{tabular}{L{2.6cm}|c|c|c|C{1.5cm}|c|C{1.3cm}}
    \hline
    % \multirow{3}{*}{} 
    \multirow{3}{*}{{\begin{tabular}[c]{@{}c@{}} \textbf{Methods} \end{tabular}}} & \multicolumn{3}{c|}{\textbf{Use data}} & \multicolumn{2}{c|}{\textbf{in Real-train}} & \multirow{3}{*}{{\begin{tabular}[c]{@{}c@{}} \# Meta \\ tasks \end{tabular}}} \\ \cline{2-6}
     & Meta-train & Meta-test & Real-train & \# samples & \begin{tabular}[c]{@{}c@{}}\# gradient \\ updates\end{tabular} &  \\ \hline
    Vanilla\_limit & - & - & \checkmark & 16 (1\%) & 100 & 0 \\ 
    Vanilla\_full & - & - & \checkmark & 960 (60\%) & 100 & 0 \\ 
    % Transfer\_fast & \checkmark & - & \checkmark & 16 (1\%) & 1 & 3 \\ 
    Transfer & \checkmark & - & \checkmark & 16 (1\%) & 100 & 3 \\ 
    MAML & \checkmark & \checkmark & \checkmark & 16 (1\%) & 1 & 3 \\ \hline
    AGILE(phase I) & \checkmark & \checkmark & \checkmark & 16 (1\%) & 1 & many \\ 
    AGILE(phase II) & \checkmark & \checkmark & \checkmark & 16 (1\%) & 1 & many \\ 
    AGILE(phase II) & \checkmark & \checkmark & \checkmark & 160 (10\%) & 1 & many \\ \hline
    \end{tabular}
    }
    \label{tab:methods}
\end{table}

\begin{table}[!t]
    \caption{Quantitative results of different methods. Original method use all available training data (60\%) and act as the upper bound while task-augmented MAML method get the highest accuracy using very few training data (1\%).}
    \centering
    \resizebox{\columnwidth}{!}{%
    % \vspace{-2mm}
    \begin{tabular}{L{3.4cm}|C{1.5cm}C{1.4cm}C{1.4cm}|C{2.7cm} C{2cm}}
    \hline
    \textbf{Methods} (Size \%) & \textbf{Precision} & \textbf{Recall} & \textbf{F1-score} & \textbf{Accuracy}($\pm$ Std) & \textbf{CI}$_{\mathbf{95}}$ \\ \hline
    Vanilla\_limit (1\%) & 0.642 & 0.622 & 0.632 & 0.637($\pm$0.062) & 0.632 - 0.642 \\
    Vanilla\_full (60\%) & 0.937 & \textbf{0.965} & \textbf{0.951} & \textbf{0.950}($\pm$0.021)  & 0.948 - 0.952 \\
    % Transfer\_fast & 0.257 & 0.262 & 0.260 & 0.253($\pm$0.026) & 0.002 \\
    Transfer (1\%) & 0.447 & 0.433 & 0.440 & 0.449($\pm$0.085) &  0.449 - 0.456 \\
    MAML (1\%) & 0.408 & 0.402 & 0.405 & 0.409($\pm$0.030) &  0.406 - 0.412 \\
    \hline
    AGILE(phase I) (1\%) & 0.791 & 0.790 & 0.791 & 0.791($\pm$0.054)  & 0.786 - 0.796 \\
    % A-TA MAML & 0.886($\pm$0.058) & 0.926($\pm$0.061) & 0.904($\pm$0.046) & 0.902($\pm$0.048)  & 0.004 \\
     AGILE(phase II) (1\%) & 0.883 & 0.926 & 0.904 & 0.902($\pm$0.048) & 0.898 - 0.906 \\
     AGILE(phase II) (10\%) & \textbf{0.950} & 0.951 & \textbf{0.951} &  \textbf{0.950}($\pm$0.044) & 0.946 - 0.954 \\ 
    \hline
    \end{tabular}
    }
    \label{tab:results} 
    % \vspace{-2mm}
\end{table}

We used three other methods for comparison. The first is the Vanilla method which is to train a network from scratch without any adaptation. We further split it into Vanilla\_limit and Vanilla\_full methods which use limited and all available training data respectively. They are acting like the lower bound and the upper bound of the performances. The second is the transfer learning model, which means the model is pre-trained on 3 meta tasks and then fine-tuned on 2 real takes. Both of these are trained with 100 gradient updates to ensure the model convergence. The third method is the plain MAML model \cite{finn2017model} with limited data and only 1 gradient update for the purpose of fast adaptation with few training data. We chose MAML as our meta-learning baseline because it achieved highly competitive performances compared to other meta-learning methods \cite{finn2017model}. All configuration differences are shown in Table \ref{tab:methods}. Note that all methods share the same neural network structure which has 4 convolution blocks. Each block consists of a convolutional layer with 32 filters of size 3*3, a batch normalization layer, and a max-pooling layer. Relu activation is used after each convolutional layer. After the last convolution block, a dense layer is added to project the 32 feature maps to only two classes. And all methods have the same learning rate $\alpha=0.01$ for the classifier. The learning rate for meta-learner is 0.001. Adam optimizer is used for all optimization. Training iterations are set as 12000 for all methods. 12 replaced tasks are used in one iteration during the training on meta-tasks. The only differences between these methods are the data and the training/adaptation framework they use.

The quantitative results on the real tasks are summarized in Table \ref{tab:results}. The upper bound we can get by using all training data is 95\% classification accuracy while the lower bound is 63.7\%. Plain transfer learning method and MAML methods are even worse than the lower bound because they were stuck in the local minimum after pre-trained on only 3 meta tasks. Our proposed task-augmented meta-learning is able to quickly adapt with few training samples and gradient updates. With only 16 training samples, the AGILE method can reach 90\% accuracy. With 160 training samples, AGILE reaches the upper bound. The 95\% confidence interval are presented in the Table \ref{tab:results}.
% We run 500 experiments so that the 95\% confidence interval are low for all methods and the results are precise.

\begin{figure}[!t]
\includegraphics[width=\textwidth]{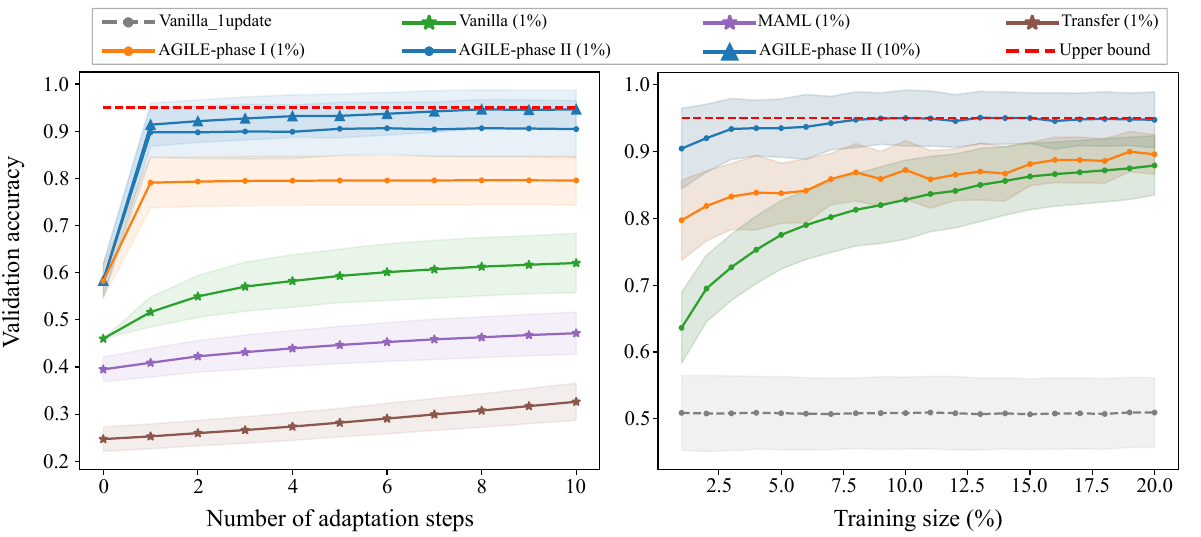}
% \vspace{-4mm}
\caption{(\textbf{a}) Comparison of different methods when adapting to new real task with only 1\% labeled training data, AGILE method shows its fast adapting ability. Upper bound is obtained by training on all 60\% training data with Vanilla method. (\textbf{b}) The impact of the training size on the AGILE method and the vanilla method.} \label{fig:adaptation}
% \vspace{-3mm}
\end{figure}

The adaptation processes are shown in Figure \ref{fig:adaptation} (\textbf{a}). The solid line is the mean value and the shaded area shows the variance. Without any prior knowledge, the original model starts with the classification accuracy around 50\%. Transferring an existing model has the lowest starting point because images labeled as ``1" in one task should be labeled as ``0'' in other tasks. The original MAML method also does not help for this brain cell type binary classification problem. Our results show that MAML is unable to adapt effectively due to the small number of training tasks. In contrast, AGILE model has a sharp increase in accuracy after 1 gradient update. Note that although AGILE model is trained for maximal performance after one gradient step, it continues to improve with additional gradient steps.

% In the second experiment, we would like to find out how many samples do we really need for solving the brain cell type classification problem because some features might only exist in certain samples. Only the proposed methods and vanilla methods are shown in the figure. Because the phase II of AGILE give us the ability to quickly extract useful features with limited samples and active learning presents the model with the most informative samples, it performs extremely well when the training size is small. One detail about the second phase of AGILE is that due to its selecting rule, the new-added training samples might not be class-balanced which is more practical compared with current meta-learning setting where all the training set is class-balanced by default. The proposed full-scale AGILE model can reach the upper bound 95\% with around 160 training samples. Meantime, the vanilla model with 100 gradient updates get the accuracy around 84\% and the vanilla model with only 1 update basically cannot learn anything. We demonstrate that with our proposed method, few training samples and little training time is enough for the new brain cell type classification problem.

Figure \ref{fig:adaptation} (\textbf{b}) illustrate the relationship between the training size and the performances of AGILE method and the baseline methods. AGILE performs extremely well when the training size is small. Since AGILE selects samples based on their uncertainties, the newly added training samples might not be class-balanced. This is more practical because difficult classes require more examples for training. Our strategy contrasts against current meta-learning setting where all the training set is class-balanced by default. AGILE method reachs validation accuracy of 95\% with 160 samples. Meantime, the vanilla model with 100 gradient updates get the accuracy around 84\% and the vanilla model with only 1 update basically cannot learn anything.

\section{Conclusion}
% \section{Conclusion and Future Work}
% \vspace{-3mm}
In this paper, we proposed a fast adaptation framework AGILE combining data augmentation, meta learning and active learning to deliver a model which is sensitive to the data distribution/task changes and able to adjust itself with few training samples and few updating steps. The results show that only 10\% of training data and 1 gradient update are enough to get the best performance on identifying unseen brain cell type. AGILE can be used in many diagnose systems or detection algorithms which have to deal with various input data. 
% The relations between the meta tasks and the real tasks requires further study in the future.

\vspace{5mm}
\noindent
\textbf{Acknowledgments.}
This work was supported by the National Science Foundation (NSF-IIS 1910973), and the Intramural Research Program of the National Institute of Neurological Disorders and Stroke, National Institutes of Health (1R01NS109118-01A1).

{\small
\bibliographystyle{splncs04}
\bibliography{ref.bib}
}




% LNCS Homepage, \url{http://www.springer.com/lncs}. Last accessed 4
% \end{thebibliography}
\end{document}